%% file: AnonymousSubmission2027.tex
\documentclass[letterpaper]{article} 
\usepackage{aaai2027}  
\usepackage[hyphens]{url}  
\usepackage{graphicx} 
\usepackage{natbib}  
\usepackage{caption} 
\usepackage{amsmath}
\usepackage{algorithm}
\usepackage{algorithmic}

\usepackage{newfloat}
\usepackage{listings}
\DeclareCaptionStyle{ruled}{labelfont=normalfont,labelsep=colon,strut=off} 
\floatstyle{ruled}
\newfloat{listing}{tb}{lst}{}
\floatname{listing}{Listing}

\usepackage{booktabs}
\usepackage{multirow}

\title{Error-Propagation Modeling for Failure Attribution\\ in LLM-Based Multi-Agent Systems}
\author{
    Jiaqi Liao\textsuperscript{\rm 1},
    Yuanzhao Zhai\textsuperscript{\rm 1},
    Huanxi Liu\textsuperscript{\rm 1},
    Xu Zhang\textsuperscript{\rm 1},\\
    Zheming Zhuang\textsuperscript{\rm 2},
    Dawei Feng\textsuperscript{\rm 1}\thanks{Co-corresponding author.},
    Bo Ding\textsuperscript{\rm 3},
    Huaimin Wang\textsuperscript{\rm 3}
}
\affiliations{
    \textsuperscript{\rm 1}College of Computer Science and Technology, National University of Defense Technology; State Key Laboratory of Complex \& Critical Software Environment, Changsha, Hunan, China\\
    \textsuperscript{\rm 2}School of Mechanical Engineering, Tianjin University, Tianjin, China\\
    \textsuperscript{\rm 3}College of Computer Science and Technology, National University of Defense Technology; National Key Laboratory of Parallel and Distributed Computing, Changsha, Hunan, China\\
}

\begin{document}

\maketitle

\begin{abstract}
\input{Content/Abstract}
\end{abstract}


\input{Content/Introduction}
\input{Content/RelatedWork}
\input{Content/ProblemFormulation.tex}
\input{Content/Method}
\input{Content/ExperimentSetup}

\input{Content/Results}

\input{Content/Conclusion_Limitation.tex}

\bibliography{aaai2027}


\end{document}


\maketitle

\section*{Overview}

This technical supplement provides material deferred from the main paper for space reasons. It is organized as follows:
\begin{itemize}[leftmargin=1.5em,itemsep=0.15em]
    \item \textbf{A.~Implementation Details:} model configuration, prompt-design principles, and output parsing.
    \item \textbf{B.~Algorithm Details:} structured trajectory construction, propagation annotation (including backward relevance tracing), and loop extraction.
    \item \textbf{C.~Additional Experiments:} inference-cost analysis.
    \item \textbf{D.~Additional Case Studies:} a qualitative Hand-Crafted example.
    \item \textbf{E.~Representative Prompts:} abbreviated key prompts used by EMFA.
\end{itemize}
The main paper is self-contained; this supplement is intended for reproducibility checks and extended reading.

\input{Content/ImplementationDetails}
\input{Content/AlgorithmDetails}
\input{Content/AdditionalResults}
\input{Content/CaseStudy}
\input{Content/RepresentativePrompts}

%% file: Content/Abstract.tex
LLM-based multi-agent systems (MASs) are increasingly used to solve complex tasks through coordinated reasoning, tool use, and interaction with external resources. However, attributing failures in such systems remains challenging because the observed outcome often does not directly reveal the error responsible for the failed execution. In this work, the attribution target is the decisive error, defined as the agent--step pair whose correction would recover the failed execution. Existing approaches largely identify suspicious steps without explicitly modeling how errors propagate across interactions or persist in unresolved loops, making decisive errors difficult to distinguish from downstream failure symptoms. We propose \textbf{E}rror-Propagation \textbf{M}odeling for \textbf{F}ailure \textbf{A}ttribution (\textbf{EMFA}). EMFA constructs a structured representation of the failed trajectory, models both cascading propagation and persistent interaction loops, and uses propagation-aware candidate screening followed by counterfactual verification to identify the decisive agent--step pair. On the Who\&When benchmark, EMFA achieves state-of-the-art step-level attribution accuracy and remains competitive at the agent level. It improves the previous best step-level results by 3.45 and 4.40 percentage points on the Hand-Crafted and Algorithm-Generated subsets, respectively.

%% file: Content/Introduction.tex
\section{Introduction}

\begin{figure}[t]
    \centering
    \includegraphics[width=\columnwidth]{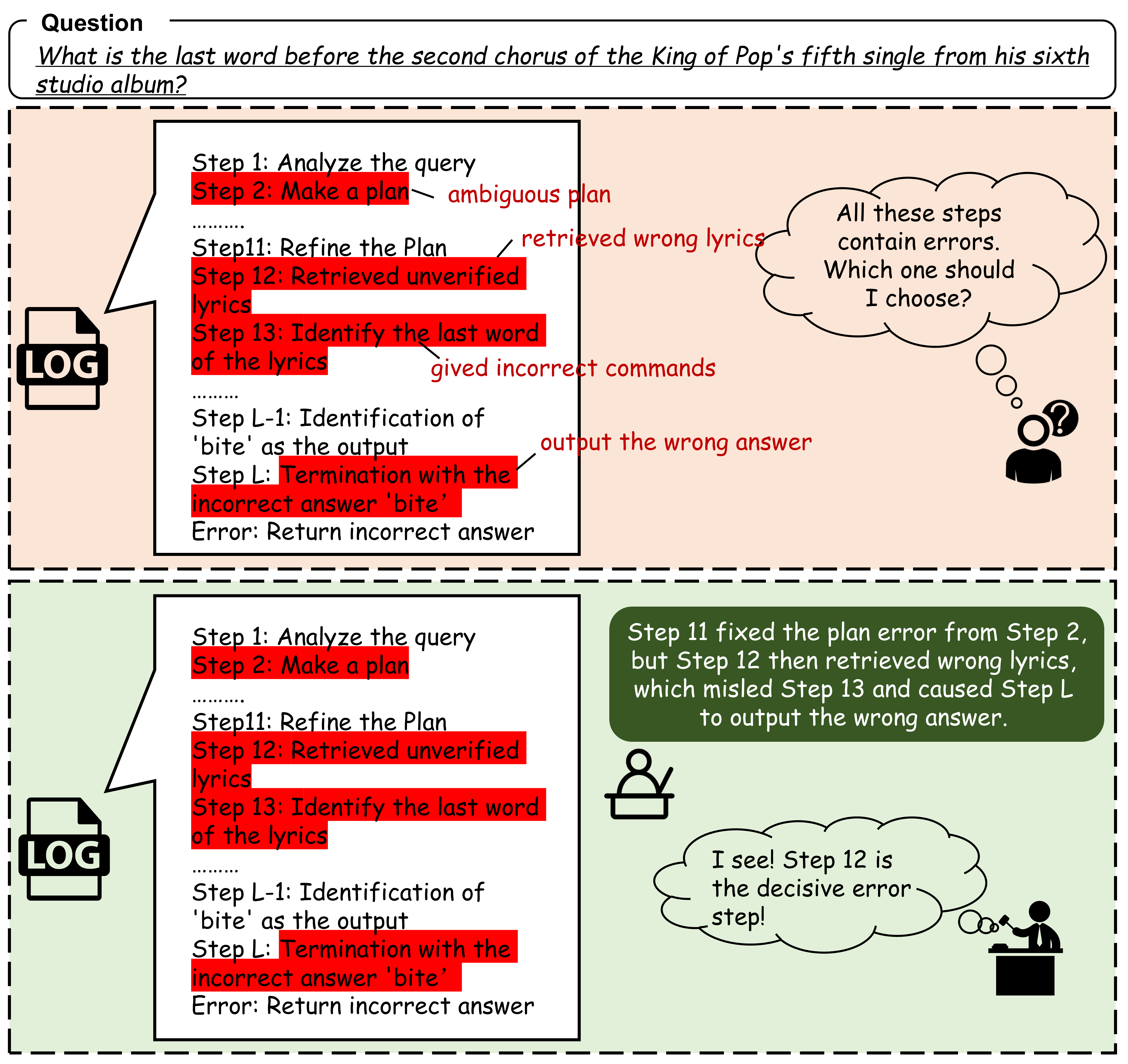}
    \caption{Illustration of why error-propagation modeling is needed to separate decisive errors from misleading failure symptoms.}
    \label{fig:why_propagation}
\end{figure}

Multi-agent systems (MASs) have emerged as a promising paradigm for solving increasingly complex tasks involving tool use, reasoning, and agent coordination by orchestrating multiple LLM-based agents to enable task decomposition, collaborative reasoning, and interaction with external resources. However, the integration of heterogeneous agents makes it difficult to trace unsuccessful executions back to the responsible agent and step. Recent empirical studies show that even advanced MAS frameworks, such as HyperAgent \citep{hyperagent}, still experience substantial failure rates (74.7\%) on the SWE-Bench Lite benchmark, with failures arising from various stages of multi-agent execution \citep{cemriWhyMultiAgentLLM2025}. These observations motivate the problem of \textbf{failure attribution}. Specifically, under the Who\&When formulation adopted in this work, the attribution target is the \textbf{decisive error}, defined as the agent-step pair whose correction would recover the failed execution \citep{whowhen}. Identifying this decisive error provides actionable signals for debugging, targeted repair, and learning from failed executions. Nevertheless, reliable attribution remains challenging because failures are rarely isolated events: errors may either propagate through dependent interactions or remain unresolved in repetitive interaction loops, eventually manifesting as misleading failure symptoms.

Existing failure attribution approaches can be broadly categorized into four paradigms: direct LLM-based methods that infer attribution targets from textual trajectories \citep{whowhen,ECHO,SDBL}; replay-based methods that estimate step responsibility through repeated trajectory executions \citep{FAMAS}; fine-tuning-based methods that learn specialized attribution models from additional supervision \citep{agentracer,graphtracer}; and structure-guided methods that exploit dependency structures to identify candidate steps \citep{chief,FALAT}. Despite their differences, these approaches largely treat attribution as selecting suspicious steps from an observed failure, without explicitly modeling how errors propagate across the trajectory or persist in unresolved interaction loops. As illustrated in Figure~\ref{fig:why_propagation}, a failed execution may contain multiple apparently erroneous steps, including deviations that are later corrected and downstream symptoms that merely inherit an earlier error. Identifying the decisive error therefore requires tracing how errors influence subsequent execution rather than judging individual steps in isolation. The figure illustrates cascading propagation across dependent steps, while failed trajectories may also contain persistent interaction loops in which agents repeatedly act without making task progress. Accordingly, we focus on these two complementary failure patterns.

Building on these observations, we propose \textbf{E}rror-\textbf{P}ropagation 
\textbf{M}odeling for \textbf{F}ailure \textbf{A}ttribution (\textbf{EMFA}), a failure attribution framework that explicitly models how errors evolve 
within failed multi-agent executions. EMFA first transforms flat execution logs into structured dependency representations to recover how information flows across agents and steps. It then models two complementary failure patterns: cascading error propagation, where an early deviation affects subsequent decisions through dependencies, and persistent interaction 
loops, where agents repeatedly act without making substantive progress. Based on these structures, EMFA constructs candidate decisive errors and 
applies counterfactual verification to distinguish truly recoverable errors from downstream carriers and failure symptoms.

On the Who\&When benchmark~\citep{whowhen}, EMFA achieves state-of-the-art step-level failure attribution accuracy, improving over the previous best by 3.45 and 4.40 percentage points on the Hand-Crafted and Algorithm-Generated subsets, respectively. It also achieves the best agent-level accuracy on the Hand-Crafted subset while remaining competitive on the Algorithm-Generated subset.

Our contributions are summarized as follows:
\begin{itemize}
    \item We introduce an error-propagation-oriented perspective on failure attribution in LLM-based multi-agent systems. This perspective explicitly models how errors propagate across dependent steps or persist in unresolved interaction loops, enabling decisive errors to be distinguished from downstream failure symptoms.

    \item We propose \textbf{EMFA}, a failure attribution framework that organizes failed trajectories into structured dependency representations, models cascading error propagation and persistent interaction loops to screen candidate steps, and applies counterfactual verification to identify the decisive agent--step pair.

    \item Experiments on the Who\&When benchmark show that \textbf{EMFA} achieves state-of-the-art step-level failure attribution accuracy while remaining competitive at the agent level.
\end{itemize}

%% file: Content/RelatedWork.tex
\section{Related Work}

\subsection{LLM-Based Multi-Agent Systems}

Recent LLM-based multi-agent systems coordinate specialized agents through either manually specified workflows or automatically optimized interaction structures. Representative workflow-based systems include AutoGen \citep{autogen}, MetaGPT \citep{metagpt}, and ChatDev \citep{chatdev}, while AFlow \citep{aflow} optimizes agent workflows or structures automatically. Despite these architectural advances, multi-agent and multi-step executions introduce complex dependencies through which errors may propagate across dependent steps or remain unresolved in repeated interactions, making the decisive error difficult to identify.

\subsection{Failure Attribution in LLM-Based Multi-Agent Systems}

The Who\&When benchmark formalizes failure attribution as identifying the decisive agent--step pair in failed execution trajectories \citep{whowhen}. Existing MAS attribution approaches can be broadly categorized into direct inference, replay-based, fine-tuning-based, and structure-guided methods. Direct LLM-based methods infer attribution targets directly from textual trajectories \citep{whowhen,ECHO,SDBL}, but often struggle to capture fine-grained dependencies. Replay-based methods such as FAMAS \citep{FAMAS} estimate responsibility through repeated executions, incurring substantial computational costs. Fine-tuning-based approaches such as AgenTracer and GraphTracer \citep{agentracer,graphtracer} learn specialized attribution models but require additional supervision and training resources \citep{kd33,kd44}.

Structure-guided methods are most closely related to EMFA. CHIEF \citep{chief} exploits hierarchical causal graphs to rank candidate errors, while FALAT \citep{FALAT} incorporates propagation analysis after candidate pruning. However, these approaches do not explicitly model complete error propagation before candidate selection, which may lead to incomplete propagation paths or confusion between decisive errors and downstream symptoms. In contrast, EMFA constructs candidate sets from explicitly modeled error-propagation structures by tracing failure-related dependencies and analyzing both cascading propagation and persistent interaction loops. It then applies counterfactual verification to identify the decisive error.

%% file: Content/ProblemFormulation.tex
\section{Problem Formulation}
\label{sec:problem}

To remain compatible with the Who\&When benchmark and follow its
established formulation, we adopt a problem setting consistent with
prior work~\citep{whowhen,chief}.

\paragraph{LLM-Based MAS.}
We consider a turn-based LLM multi-agent system in which exactly one
agent acts at each execution step. Let $\mathcal{N}$ denote the set of
agents. Given a task query $Q$, the system produces an execution
trajectory
\begin{equation}
    \tau=(x_1,x_2,\ldots,x_L),
    \qquad
    x_i=(a_i,u_i),
\end{equation}
where $i\in\{1,\ldots,L\}$ is the step index,
$a_i\in\mathcal{N}$ is the agent acting at step $i$, and $u_i$ denotes
the action, observation, reasoning, or result produced at that step.
Let $Z(\tau)\in\{0,1\}$ denote the task outcome, where $Z(\tau)=1$
indicates failure and $Z(\tau)=0$ indicates success.

\paragraph{Final Failure Symptom.}
For a failed trajectory $\tau$, let $i_{\mathrm{final}}$ denote the
index of the final failure step
$x_{\mathrm{final}}=x_{i_{\mathrm{final}}}$, at which the unsuccessful
outcome becomes observable, e.g., through an incorrect final answer,
premature termination, or an unresolved task requirement. We refer to
the observable task-level mismatch at $x_{\mathrm{final}}$ as the
\emph{final failure symptom}. This step need not coincide with the
decisive step, because the observed symptom may be a downstream
consequence of an earlier error.

\paragraph{Failure Attribution.}
Given a failed trajectory $\tau$ with $Z(\tau)=1$, failure attribution
aims to identify the \emph{decisive agent--step pair}. For each step
$i$ and its acting agent $a_i$, let
$\widetilde{\tau}^{(a_i,i)}$ denote the counterfactual execution
obtained by correcting the action at step $i$ and allowing the
subsequent execution to unfold accordingly. We define the
decisive-error indicator as
\begin{equation}
\Delta_{a_i,i}(\tau)=
\begin{cases}
1, & \text{if } Z\!\left(\widetilde{\tau}^{(a_i,i)}\right)=0,\\
0, & \text{otherwise}.
\end{cases}
\end{equation}
Thus, $(a_i,i)$ is a \emph{decisive error} if correcting the action at
step $i$ would recover the failed execution.

A failed trajectory may contain multiple agent--step pairs satisfying
this counterfactual criterion. Following the benchmark annotation
protocol, we take the earliest decisive error as the attribution
target:
\begin{equation}
    i^*
    =
    \min\left\{
        i \,\middle|\,
        \Delta_{a_i,i}(\tau)=1
    \right\},
    \qquad
    a^*=a_{i^*}.
\end{equation}
Here, $i^*$ is the index of the decisive step, $a^*$ is the decisive
agent, and $x_{i^*}$ is the corresponding complete execution step.
The objective is therefore to construct an attribution function
\begin{equation}
    F(Q,\tau)=(a^*,i^*),
\end{equation}
that recovers the ground-truth decisive agent--step pair from the task
query and the observed failed trajectory.

%% file: Content/Method.tex
\section{Method}

\begin{figure*}[t]
\centering
\includegraphics[width=\textwidth]{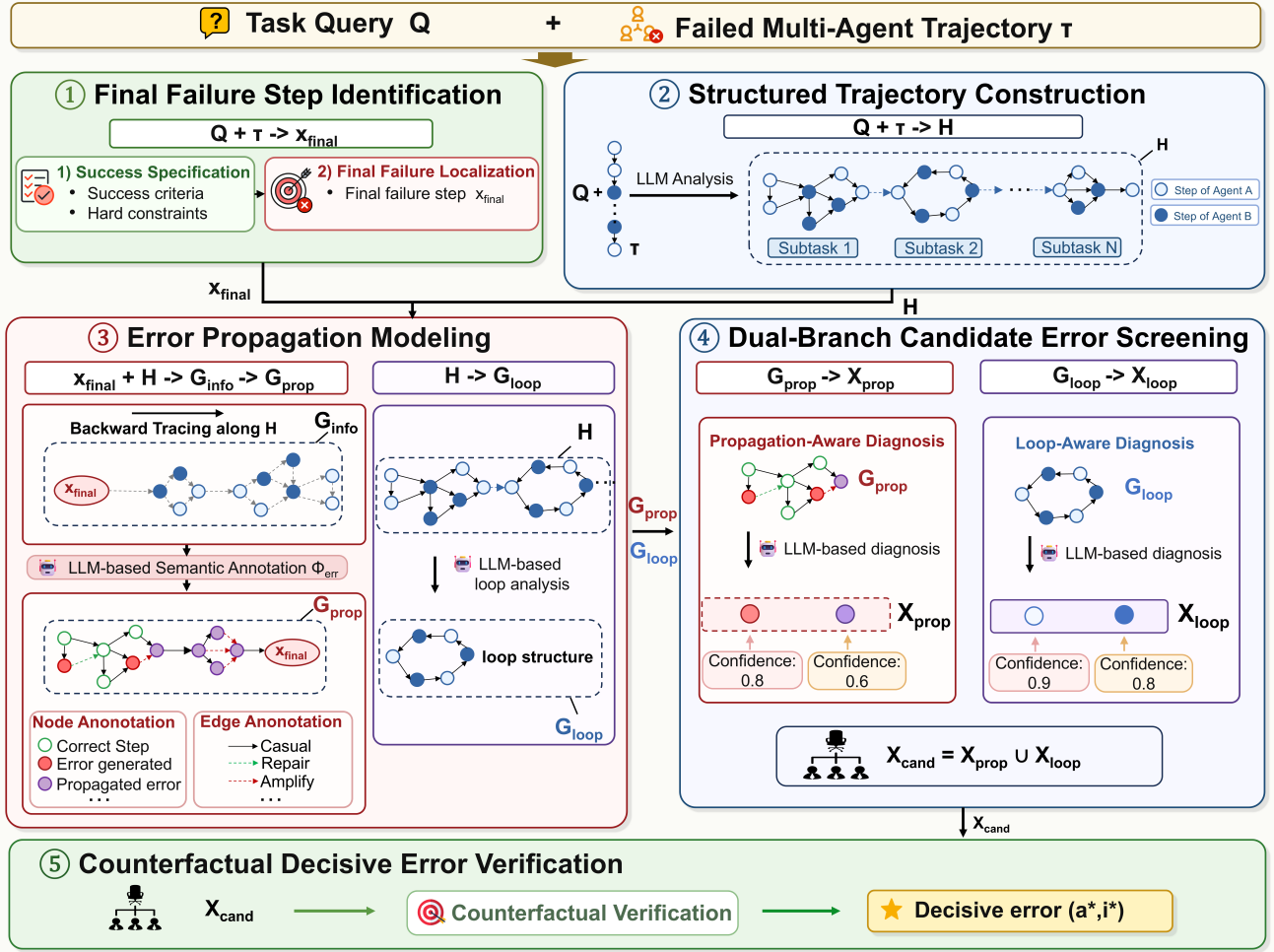}
\caption{
Overview of EMFA. Given a query $Q$ and a failed trajectory $\tau$, EMFA locates the final failure step corresponding to the final failure symptom and structures the trajectory. It then traces failure-related information backward, reconstructs how errors propagate chronologically using an LLM, screens candidate errors through propagation and loop branches, and applies counterfactual decisive error verification to select the decisive error step $x^{*}$ and its responsible agent $a^{*}$.
}
\label{fig:emfa_overview}
\end{figure*}

Our method, \textbf{E}rror-Propagation \textbf{M}odeling for \textbf{F}ailure \textbf{A}ttribution (\textbf{EMFA}), identifies the decisive error by tracing how the final failure symptom arises from earlier errors in a failed multi-agent execution. As illustrated in Figure~\ref{fig:emfa_overview}, EMFA consists of five stages. First, \emph{Final Failure Step Identification} locates the step at which the failure symptom is exposed. Second, \emph{Structured Trajectory Construction} converts the flat execution log into a structured representation of agents, steps, and their dependencies. Third, \emph{Error Propagation Modeling} captures both error propagation across dependent steps and unresolved errors that persist through repetitive interaction loops. Fourth, \emph{Dual-Branch Candidate Screening} extracts candidate errors from the propagation and loop structures. Finally, \emph{Counterfactual Decisive Error Verification} evaluates these candidates and identifies the decisive agent--step pair.

\subsection{Final Failure Step Identification}
\label{sec:final_failure}

Errors may arise early in a multi-agent execution but become observable only when the system produces an incorrect answer or terminates without resolving the task. EMFA therefore first identifies the \emph{final failure step}, denoted as $x_{\mathrm{final}}$, at which the final failure symptom is exposed. This step serves as the terminal anchor for subsequent dependency tracing.

EMFA first extracts a task-success specification $C_{\mathrm{success}}$ from the query $Q$. The specification describes the expected outcome and any explicit requirements, such as source, format, temporal, or numerical constraints. It is derived solely from the query so that the evaluation criteria remain independent of the failed execution.

For a trajectory that produces an explicit final answer, EMFA scans the last $w$ steps in reverse order and selects the latest agent step that directly answers the query, excluding intermediate plans, tool outputs, and progress updates. We set $w=5$ to cover the terminal response region. If the trajectory reaches the maximum execution length without producing an explicit answer, its last step is selected instead, as it marks where the unresolved execution terminates. The selected step is evaluated against $C_{\mathrm{success}}$ to characterize the final failure symptom.

The final failure step is used only as an observation anchor. It may be a downstream manifestation of an earlier error and therefore does not necessarily coincide with either the decisive error or a propagation source.

\subsection{Structured Trajectory Construction}
\label{sec:trajectory_structuring}

A raw execution trajectory records the temporal order of agent actions but does not explicitly reveal how information is produced, transferred, and reused across agents and subtasks. EMFA therefore converts the flat trajectory into a two-level dependency representation
\begin{equation}
\mathcal{H}
=
\left(
\mathcal{X},
\mathcal{S},
\mathcal{E}_{\mathrm{step}},
\mathcal{E}_{\mathrm{sub}}
\right),
\end{equation}
where $\mathcal{X}$ contains the trajectory steps together with their agent identities and subtask assignments, and $\mathcal{S}=\{S_1,\ldots,S_M\}$ is a sequence of contiguous, non-overlapping semantic subtasks.

Within each subtask, $\mathcal{E}_{\mathrm{step}}$ captures directed producer--consumer information dependencies between execution steps, with particular emphasis on information transferred across agents. An edge $(x_i,x_j)\in\mathcal{E}_{\mathrm{step}}$ indicates that information produced at $x_i$ is subsequently consumed at $x_j$, and records the information transferred between them. While most such edges represent inter-agent communication, dependencies between steps of the same agent are retained when they are necessary to reconstruct how task-relevant information is carried forward.

Across subtasks, $\mathcal{E}_{\mathrm{sub}}$ connects an earlier subtask to a later one that reuses its outputs, including dependencies between non-adjacent subtasks. The subtask-level structure provides a coarse tracing scope, while the step-level dependencies support fine-grained producer identification during subsequent backward tracing.

We additionally apply deterministic post-processing to recover explicit local interactions that may be omitted during semantic extraction. When an instruction is immediately followed by a response from the addressed agent, EMFA adds the corresponding instruction--response dependency if it is absent. This rule completes clear interaction links without treating temporal adjacency alone as evidence of information dependency.

\subsection{Error Propagation Modeling}
\label{sec:error_propagation}

The structured trajectory $\mathcal{H}$ captures broad execution dependencies, many of which may be unrelated to the final failure symptom. EMFA therefore extracts two complementary failure-related structures: an error-propagation graph $\mathcal{G}_{\mathrm{prop}}$, which describes how a relevant deviation propagates along information dependencies, and a persistent-loop graph $\mathcal{G}_{\mathrm{loop}}$, which captures unresolved repetitive interactions.

\paragraph{Backward Relevance Tracing.}
Starting from the final failure step $x_{\mathrm{final}}$, EMFA
traverses the dependency structure in $\mathcal{H}$ backward to collect
upstream subtasks and execution steps whose outputs are directly or
indirectly reused by the current step. The reached steps and their
producer--consumer dependencies form the relevant information graph
\begin{equation}
\mathcal{G}_{\mathrm{info}}
=
\left(
\mathcal{V}_{\mathrm{info}},
\mathcal{E}_{\mathrm{info}}
\right).
\end{equation}
Although $\mathcal{G}_{\mathrm{info}}$ is constructed through backward
traversal, its edges preserve the actual producer-to-consumer direction.
Thus, an edge $x_i\!\rightarrow\!x_j$ indicates that information
produced at $x_i$ is subsequently used at $x_j$. This process isolates
the information channels potentially related to the observed failure.
The detailed backward relevance tracing algorithm is provided in the
technical supplement.

\paragraph{Forward Propagation Annotation.}
EMFA then processes the steps in $\mathcal{G}_{\mathrm{info}}$ in chronological order and uses an LLM to annotate how task-relevant information changes along its nodes and edges. The annotations record whether a deviation is introduced, propagated, transformed, or corrected, thereby distinguishing propagation sources, propagation carriers, recovered deviations, unrecovered deviations, and the failure symptom exposed at $x_{\mathrm{final}}$. The resulting graph is
\begin{equation}
\mathcal{G}_{\mathrm{prop}}
=
\left(
\mathcal{G}_{\mathrm{info}},
\Phi_{\mathrm{err}}
\right),
\end{equation}
where $\Phi_{\mathrm{err}}$ denotes the semantic propagation annotations. Simple deterministic checks for explicit or omitted required actions complement the LLM annotations. A propagation source describes the origin of a reconstructed propagation path and does not necessarily coincide with the decisive error.

\paragraph{Persistent Loop Extraction.}
Error propagation along producer--consumer dependencies does not capture failures in which agents repeatedly pursue the same unresolved objective without making substantive progress. EMFA therefore uses an LLM to identify persistent interaction loops directly from $\mathcal{H}$ by comparing the objectives, actions, observations, and progress of repeated interaction rounds. The resulting $\mathcal{G}_{\mathrm{loop}}$ connects steps that repeatedly address the same unresolved objective without resolving it. Because such loops may span multiple subtasks and need not follow an ordinary information-dependency path, they are extracted independently of $\mathcal{G}_{\mathrm{info}}$.

Together, $\mathcal{G}_{\mathrm{prop}}$ and $\mathcal{G}_{\mathrm{loop}}$ provide complementary evidence for subsequent candidate screening.

\subsection{Dual-Branch Candidate Error Screening}

EMFA independently screens candidate errors from the error-propagation and persistent-loop structures, retaining complementary causal explanations for subsequent counterfactual decisive error verification.

\paragraph{Propagation-Branch Screening.}
Given $\mathcal{G}_{\mathrm{prop}}$, an LLM selector compares the reconstructed propagation events and returns a propagation candidate set $\mathcal{X}_{\mathrm{prop}}$ containing at most two agent--step pairs. It prioritizes the earliest unrecovered deviation that materially affects subsequent reasoning or actions. Unless a later incorrect step introduces a new error, substantially transforms an existing one, or makes a consequential control decision, it is treated as a propagation carrier, or as the failure symptom when it coincides with $x_{\mathrm{final}}$. Rule-based checks further clarify recurring responsibility boundaries, including plan author versus executor, faulty transformation versus raw evidence, and premature completion versus an earlier recoverable limitation.

\paragraph{Loop-Branch Screening.}
Each loop in $\mathcal{G}_{\mathrm{loop}}$ is analyzed independently to identify the earliest step that initiates or sustains the unresolved repetition. The analysis distinguishes, when applicable, between an ambiguous or incorrect instruction and a failure to follow or correctly execute a valid instruction, while also considering omitted actions and failures to adapt after repeated unsuccessful attempts. When the same unresolved error recurs across multiple rounds, its first occurrence is selected as the group-level candidate. An aggregate selector retains at most two loop candidates in $\mathcal{X}_{\mathrm{loop}}$.

The final candidate pool is the deduplicated union
\begin{equation}
X_{\mathrm{cand}}=X_{\mathrm{prop}}\cup X_{\mathrm{loop}}.
\end{equation}
Each branch contributes at most two candidates, so full EMFA
evaluates up to four candidates, while each single-branch
variant evaluates up to two.

\subsection{Counterfactual Decisive Error Verification}

EMFA performs model-based counterfactual assessment rather
than environment replay, consistent with the expert-judgment
annotation protocol of Who\&When. For each candidate
$(a_i,i)\in X_{\mathrm{cand}}$, the verifier assesses whether minimally repairing the step would prevent the downstream failure and satisfy $C_{\mathrm{success}}$. A candidate is considered more causally decisive when its repair blocks the downstream failure process rather than merely correcting a later propagation carrier or the final symptom.

When multiple candidate repairs appear capable of recovering the execution, EMFA resolves the remaining ambiguity using predefined responsibility-boundary principles that distinguish the step responsible for introducing an error from downstream steps that merely execute, propagate, or expose it. Integrating counterfactual recovery with these principles, the verifier returns
\begin{equation}
(\hat{a},\hat{i})
=
F_{\mathrm{cf}}
\left(
Q,\tau,C_{\mathrm{success}},
\mathcal{X}_{\mathrm{cand}}
\right),
\end{equation}
where $(\hat{a},\hat{i})$ is EMFA's predicted decisive agent--step pair. Here, $\hat{i}$ denotes the predicted index of the earliest decisive step, and $\hat{a}$ is the agent acting at that step. The prediction is evaluated against the ground-truth pair $(a^*,i^*)$.

%% file: Content/ExperimentSetup.tex
\section{Experiment Setup}

\subsection{Dataset}

We evaluate EMFA on the \textit{Who\&When} benchmark \citep{whowhen}, which annotates the decisive error step (``When'') and responsible agent (``Who'') for failed multi-agent trajectories. The test set contains two subsets: (1)~\textbf{Hand-Crafted}, consisting of 58 logs from Magentic-One \citep{magenticone}, and (2)~\textbf{Algorithm-Generated}, consisting of 126 logs generated from diverse architectures by CaptainAgent \citep{captainagent}. Tasks are derived from GAIA \citep{gaia} and AssistantBench \citep{assistantbench}. Each sample provides an execution trajectory with agent identifiers and step-level outputs, together with annotated failure steps, responsible agents, and failure reasons. Ground-truth annotations are obtained through multi-round expert consensus.

\subsection{Baselines}

Given the emerging nature of MAS failure attribution, we compare EMFA against ten representative approaches spanning four paradigms.
\textit{\textbf{Random}} randomly selects an agent--step pair as a lower-bound baseline.
\textit{\textbf{LLM-based Prompting}} includes All-at-Once, Step-by-Step, and Binary Search from Who\&When \citep{whowhen}, together with ECHO \citep{ECHO} and SDBL \citep{SDBL}, which directly infer or narrow attribution targets from execution trajectories.
\textit{\textbf{Fine-tuning-based Methods}} include AgenTracer and GraphTracer \citep{agentracer,graphtracer}, which learn attribution models from additional supervision.
\textit{\textbf{Structure-guided Methods}} include CHIEF \citep{chief} and FALAT \citep{FALAT}, which exploit trajectory structures or propagation information for attribution.
We exclude FAMAS \citep{FAMAS} because it requires access to task ground truth $\mathcal{G}$, whereas all compared methods operate without $\mathcal{G}$.

\subsection{Evaluation Metrics}
\input{Content/table/main_results.tex}
Following prior work \citep{whowhen,agentracer}, we employ two evaluation metrics: \textbf{agent-level accuracy} and \textbf{step-level accuracy}. Agent-level accuracy measures whether a method correctly identifies the responsible failure agent, while step-level accuracy measures whether it correctly identifies the exact decisive error step. All results use a strict Top-1 criterion. EMFA results are averaged over three independent runs, while published baseline values follow the reporting protocols of their original papers. While the ground-truth answer to each query is provided in the benchmark, real-world failure attribution often lacks access to such answers. Thus, unless otherwise specified, all experiments in this paper are conducted without using task ground truth $\mathcal{G}$.

\subsection{Implementation Details}

We implement EMFA in Python~3.11 on Ubuntu. Unless otherwise specified, the base LLM is GPT-4o, with API temperature set to~$0$, a maximum generation length of $4{,}096$ tokens. For baselines without complete public implementations, we report published results when they exceed our applicable GPT-4o reproductions, thereby avoiding underestimation of baseline performance. Additionally, detailed prompts, configurations, and further cost and qualitative analyses are provided in the supplementary materials.

%% file: Content/table/main_results.tex
\begin{table*}[!htbp]
    \centering
    \setlength{\tabcolsep}{5pt}
    \caption{Main results on the Who\&When benchmark. Each cell reports accuracy (\%). Best results in each column are in \textbf{bold}; second-best results are underlined. For EMFA, values in parentheses indicate the gap to the second-best result when EMFA ranks first, or to the best result otherwise.}
    \label{tab:main_results}
    \begin{tabular}{llcccc}
    \toprule
    \multirow{2}{*}{\textbf{Type}} & \multirow{2}{*}{\textbf{Method}} & \multicolumn{2}{c}{\textbf{Hand-Crafted Dataset}} & \multicolumn{2}{c}{\textbf{Algorithm-Generated Dataset}} \\
    \cmidrule(lr){3-4} \cmidrule(lr){5-6}
     & & Agent\_ACC & Step\_ACC & Agent\_ACC & Step\_ACC \\
    \midrule
    Heuristic & Random & 12.00 & 4.20 & 29.10 & 19.10 \\
    \midrule
    \multirow{5}{*}{\shortstack[l]{LLM-based\\Prompting}}
     & All-at-Once & 48.28 & 5.17 & 59.52 & 15.87 \\
     & Step-by-Step & 34.30 & 6.90 & 28.30 & 17.80 \\
     & Binary Search & 36.20 & 6.90 & 30.10 & 16.60 \\
     & ECHO & 67.90 & 26.80 & 67.20 & 27.20 \\
     & SDBL & 68.52 & 27.78 & 63.49 & 39.68 \\
    \midrule
    \multirow{2}{*}{\shortstack[l]{Fine-tuning-\\based}}
     & AgenTracer & 63.82 & 20.68 & 63.73 & 37.30 \\
     & GraphTracer & 69.74 & 27.97 & \underline{67.42} & 44.35 \\
    \midrule
    \multirow{2}{*}{\shortstack[l]{Structure-guided}}
     & CHIEF & 72.41 & \underline{29.31} & \textbf{68.80} & \underline{45.60} \\
     & FALAT & \underline{73.00} & 29.10 & 59.70 & 38.00 \\
    \midrule
     & \textbf{EMFA (Ours)} & \textbf{74.14} (+1.14) & \textbf{32.76} (+3.45) & 66.67 (-2.13) & \textbf{50.00} (+4.40) \\
    \bottomrule
    \end{tabular}
\end{table*}

%% file: Content/Results.tex
\section{Results}

\subsection{Main Results}

As shown in Table~\ref{tab:main_results}, EMFA achieves the best
step-level attribution accuracy on both subsets of the Who\&When
benchmark. It improves the previous best results, obtained by CHIEF,
by 3.45 and 4.40 percentage points on the Hand-Crafted and
Algorithm-Generated subsets, respectively, while substantially
outperforming direct prompting baselines. Across all methods,
step-level accuracy is markedly lower than agent-level accuracy,
highlighting the greater difficulty of precisely locating the decisive step. EMFA's consistent gains on this stricter metric across both subsets demonstrate its advantage in fine-grained failure attribution. At the agent level, EMFA achieves the best result on the Hand-Crafted subset and remains competitive on the Algorithm-Generated subset.

\subsection{Effect of Trajectory Length on Failure Attribution}
\begin{figure}[t]
    \centering
    \includegraphics[width=\columnwidth]{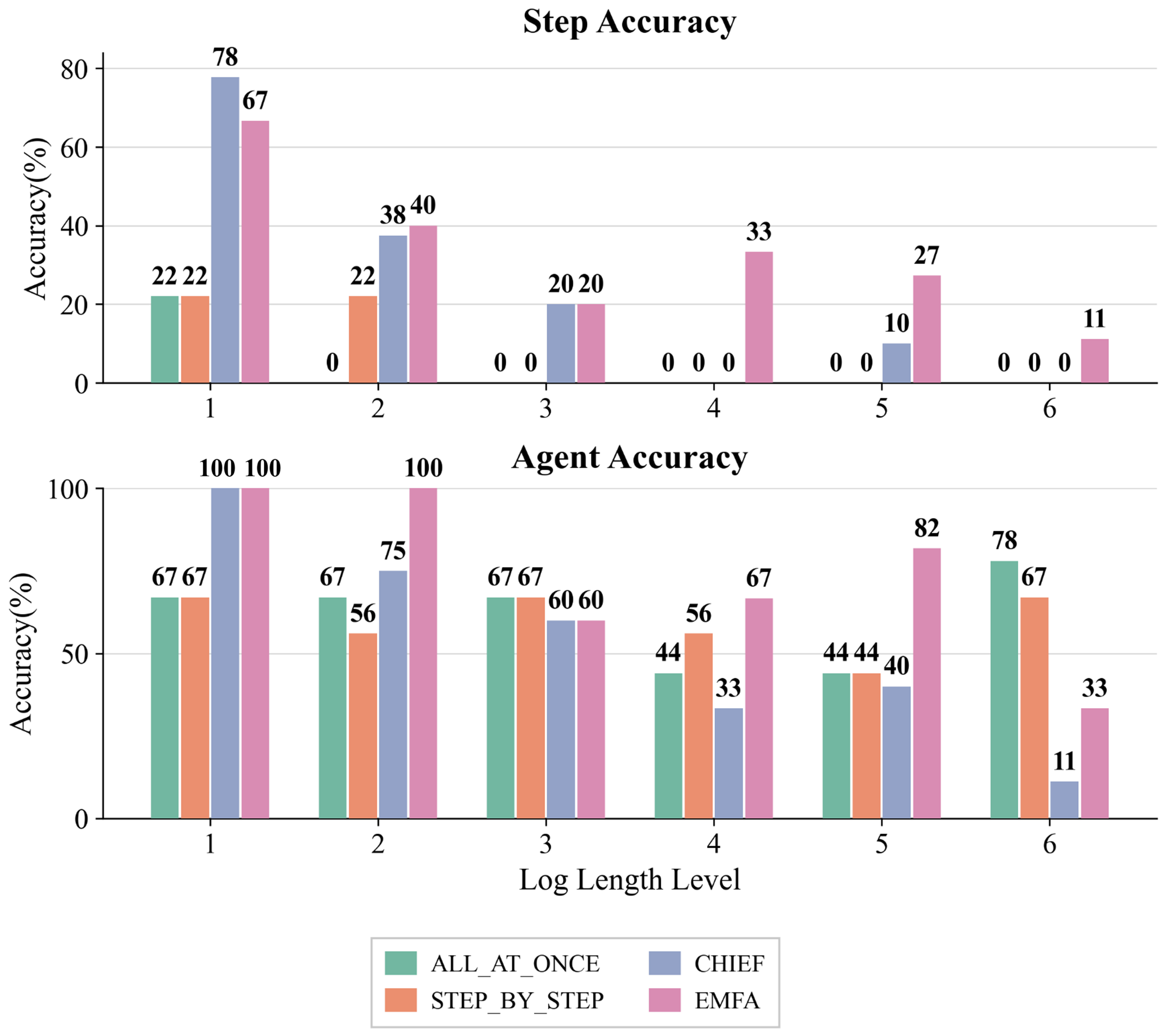}
    \caption{Agent- and step-level accuracy of All-at-Once, Step-by-Step, CHIEF (SOTA), and EMFA across six log-length tiers on the Hand-Crafted subset.}
    \label{fig:log_length}
\end{figure}

\input{Content/table/llm_backbone.tex}

\input{Content/table/ablation.tex}
Following the protocol of \citet{SDBL}, we evaluate agent and step accuracy across different log lengths on the \textbf{Hand-Crafted} subset, as longer trajectories typically increase attribution ambiguity by enlarging the distance between decisive errors and final failure symptoms. Test samples are grouped into six tiers by step count: Level~1 (5--16 steps, $n{=}9$), Level~2 (17--21, $n{=}10$), Level~3 (24--32, $n{=}10$), Level~4 (33--53, $n{=}9$), Level~5 (59--106, $n{=}11$), and Level~6 (113--130, $n{=}9$). We compare All-at-Once, Step-by-Step, the previous SOTA structure-guided method CHIEF, and EMFA under this stratification; CHIEF results are from our reproduction with Claude-Opus-4.6-thinking.

Performance across different trajectory lengths reveals a trade-off between fine-grained localization and agent-level attribution. EMFA maintains relatively strong step-level accuracy across the six tiers, suggesting that propagation-aware reasoning remains effective on longer execution traces. However, EMFA and CHIEF show lower agent-level accuracy on some longer trajectories, one possible explanation is that structure-guided methods introduce stronger inductive biases and may miss failure
patterns not represented by their structures: EMFA focuses on propagation and persistent interaction patterns, while CHIEF relies on predefined agent dependencies. Such biases may occasionally miss responsible agents when failures arise from unmodeled failure propagation patterns. Nevertheless, step-level attribution remains the more fine-grained objective, and EMFA's consistent advantage highlights the effectiveness of explicit error propagation modeling. Future work will explore additional failure propagation patterns to further improve attribution robustness.

\subsection{Further Analysis}
We conduct further analyses to understand the robustness and effectiveness of EMFA beyond the main benchmark results, including backbone robustness and ablation studies. Additional qualitative analysis and inference-cost analysis are provided in the technical supplement.

\paragraph{Performance on Different LLMs.}
As shown in Table~\ref{tab:llm_backbone}, EMFA remains effective across all tested backbones, although its absolute performance varies with the underlying LLM. GPT-4o achieves the best results on both subsets, with a particularly clear advantage on Algorithm-Generated step-level attribution (50.00\% versus 39.68--40.48\% for the open-weight models). Among the open-weight backbones, Qwen3.5-397B-A17B performs best on the Hand-Crafted subset, while the three models achieve comparable step-level accuracy on the Algorithm-Generated subset. Overall, these results indicate that EMFA can be instantiated with different base LLMs, while backbone selection continues to affect attribution performance.

\paragraph{Ablation Study.}
We ablate the dual-branch candidate-screening stage while
keeping Counterfactual Decisive Error Verification fixed. Each retained branch contributes at most two candidates. As shown in Table~\ref{tab:ablation}, full EMFA consistently outperforms direct LLM screening and both single-branch variants across all evaluation metrics.
Removing the Propagation Branch causes a substantial performance degradation, particularly in step-level accuracy,
showing that propagation-aware candidates contribute strongly
to the full system. In contrast, removing the Loop Branch results in a smaller drop on the Hand-Crafted subset but causes a pronounced degradation on the Algorithm-Generated subset, where step-level accuracy decreases from 50.00\% to 33.33\%. This indicates that loop-aware screening provides complementary benefits by identifying failures caused by persistent unproductive interactions.

Overall, the full dual-branch design provides the strongest end-to-end attribution performance, indicating complementary candidate coverage from propagation and loop analysis.

%% file: Content/table/llm_backbone.tex
\begin{table}[t]
    \centering
    \setlength{\tabcolsep}{3.5pt}
    \caption{
    EMFA performance on the Who\&When benchmark using various backbone LLMs. Agent and Step denote Agent\_ACC and Step\_ACC, respectively. Best results in each column are in \textbf{bold}.
    }
    \label{tab:llm_backbone}
    \begin{tabular}{@{}lcccc@{}}
    \toprule
    \multirow{2}{*}{\textbf{LLM Backbones}} & \multicolumn{2}{c}{\textbf{Hand-Crafted}} & \multicolumn{2}{c}{\textbf{Alg.-Generated}} \\
    \cmidrule(lr){2-3} \cmidrule(lr){4-5}
     & Agent   & Step   & Agent   & Step  \\
    \midrule
    DeepSeek-V4-Flash & 61.40 & 24.56 & 55.56 & 40.48 \\
    Qwen3.5-397B-A17B & 72.41 & 29.31 & 56.35 & 39.68 \\
    GLM-4.7 & 67.24 & 25.86 & 53.59 & 40.48 \\
    \textbf{GPT-4o} & \textbf{74.14} & \textbf{32.76} & \textbf{66.67} & \textbf{50.00} \\
    \bottomrule
    \end{tabular}
\end{table}

%% file: Content/table/ablation.tex
\begin{table}[t]
    \centering
    \setlength{\tabcolsep}{3.5pt}
    \caption{Ablation study of dual-branch candidate construction in EMFA. Each active branch contributes at most two candidates, while counterfactual verification remains unchanged.}
    \label{tab:ablation}
    \begin{tabular}{@{}lcccc@{}}
    \toprule
    \multirow{2}{*}{\textbf{EMFA Variants}} & \multicolumn{2}{c}{\textbf{Hand-Crafted}} & \multicolumn{2}{c}{\textbf{Alg.-Generated}} \\
    \cmidrule(lr){2-3} \cmidrule(lr){4-5}
     & Agent  & Step  & Agent  & Step  \\
    \midrule
    Direct LLM Screening & 53.45 & 18.97 & 42.06 & 29.35 \\
    w/o Propagation Branch & 27.59 & 1.72 & 21.43 & 13.49 \\
    w/o Loop Branch & 55.17 & 31.03 & 44.44 & 33.33 \\
    \textbf{EMFA} & \textbf{74.14} & \textbf{32.76} & \textbf{66.67} & \textbf{50.00} \\
    \bottomrule
    \end{tabular}
\end{table}

%% file: Content/Conclusion_Limitation.tex
\section{Conclusion and Limitations}

This paper presents EMFA, a failure attribution framework for LLM-based multi-agent systems that models cascading error propagation and persistent interaction loops to identify decisive errors. Rather than localizing observable failure symptoms, EMFA traces failure-related dependencies, models error propagation through propagation and persistent interaction patterns, and applies counterfactual verification to identify decisive agent--step pairs. Experiments on the Who\&When benchmark demonstrate that EMFA achieves state-of-the-art step-level attribution accuracy while remaining competitive at the agent level.

Despite its effectiveness, EMFA introduces additional inference-time overhead due to explicit trajectory structuring and propagation modeling. Future work will explore more efficient propagation modeling strategies to reduce computational costs while preserving attribution reliability.

%% file: Supplement/Content/ImplementationDetails.tex
\section{A.~Implementation Details}

\subsection{A.1~Model Configuration}

Unless otherwise specified, EMFA uses GPT-4o as the backbone LLM for all stages, with temperature set to~$0$ and a maximum generation length of $4{,}096$ tokens. Experiments are run in Python~3.11. The public release defaults to the official OpenAI Chat Completions API; alternative OpenAI-compatible backends (e.g., DeepSeek) are supported through configuration. Stage-level parallelism processes multiple samples concurrently, but LLM calls within a sample follow the stage dependency order described in the main paper.

For baselines without complete public implementations, we report published results when they exceed our GPT-4o reproductions under the same protocol, so as not to underestimate baseline performance. EMFA results in the main paper are averaged over three independent runs.

\subsection{A.2~Prompt Design Overview}

EMFA uses stage-specific prompts with a shared design pattern: (i)~a short system instruction that constrains the task and requires strict JSON output; (ii)~a user prompt that supplies only the evidence needed for that stage; and (iii)~explicit decision rules that discourage solving the original user query. Prompts never reveal ground-truth mistake annotations. Across stages:

\begin{itemize}[leftmargin=1.5em,itemsep=0.15em]
    \item \textbf{Final Failure Step Identification} extracts a task-success specification $C_{\mathrm{success}}$ from the query alone, then localizes $x_{\mathrm{final}}$ by reverse-scanning the terminal trajectory window.
    \item \textbf{Structured Trajectory Construction} asks the model to partition the log into contiguous subtasks and recover producer--consumer dependencies.
    \item \textbf{Error Propagation Modeling} provides reconstructed graph context rather than the full raw log whenever possible, reducing distraction from irrelevant steps.
    \item \textbf{Dual-Branch Candidate Screening} selects at most two candidates from each of the propagation and loop branches to form $\mathcal{X}_{\mathrm{cand}}$.
    \item \textbf{Counterfactual Verification} evaluates candidates in $\mathcal{X}_{\mathrm{cand}}$ under a minimally repaired counterfactual, together with responsibility-boundary principles.
\end{itemize}
Representative prompt fragments appear in Section~E.

\subsection{A.3~Output Parsing}

All LLM stages request strict JSON. The implementation (i)~strips optional Markdown fences, (ii)~attempts direct \texttt{json.loads}, and (iii)~falls back to extracting the outermost brace-delimited object when needed. Missing or malformed fields trigger stage-local retries (e.g., regenerating $C_{\mathrm{success}}$) or conservative defaults (e.g., falling back to the last trajectory step when no explicit answer step is found). Deterministic post-processors then normalize step indices, agent names, and graph edges before writing per-sample JSON artifacts under each stage directory.

%% file: Supplement/Content/AlgorithmDetails.tex
\section{B.~Algorithm Details}

This section expands the Method section of the main paper. We detail structured trajectory construction, the backward relevance tracing procedure used to form $\mathcal{G}_{\mathrm{info}}$, the propagation annotation schema for $\mathcal{G}_{\mathrm{prop}}$, and persistent-loop extraction for $\mathcal{G}_{\mathrm{loop}}$.

\subsection{B.1~Structured Trajectory Construction}

A raw multi-agent log records only the temporal order of messages. EMFA converts it into the two-level dependency representation
\begin{equation}
\mathcal{H}
=
\left(
\mathcal{X},
\mathcal{S},
\mathcal{E}_{\mathrm{step}},
\mathcal{E}_{\mathrm{sub}}
\right)
\end{equation}
used throughout later stages. Construction proceeds as follows.

\paragraph{Semantic subtask decomposition.}
Given query $Q$ and trajectory $\tau$, an LLM partitions the execution into a sequence of contiguous, non-overlapping semantic subtasks $\mathcal{S}=\{S_1,\ldots,S_M\}$. Each subtask is associated with a step range, a short description, and lightweight evidence supporting the cut.

\paragraph{Step- and subtask-level dependencies.}
Within each subtask, the model (and a deterministic post-processing rule) recover directed producer--consumer edges $\mathcal{E}_{\mathrm{step}}$, with emphasis on inter-agent information transfer. Across subtasks, $\mathcal{E}_{\mathrm{sub}}$ records reuse of earlier outputs by later subtasks, including non-adjacent dependencies when needed. The resulting $\mathcal{H}$ provides a coarse subtask scope for tracing and a fine-grained step graph for producer identification. Temporal adjacency alone is never treated as sufficient evidence of information dependency.

\subsection{B.2~Propagation Annotation Schema}

Error-propagation modeling first isolates failure-related information channels, then annotates how deviations move along those channels.

\paragraph{Backward relevance tracing.}
EMFA initializes an anchor queue with the final failure step $x_{\mathrm{final}}$. For each anchor $x_j$, it follows intra-subtask dependencies and cross-subtask data-transfer links in $\mathcal{H}$ to identify earlier steps that directly inform the anchor. When multiple cross-subtask producers are associated with the same data item, an LLM filter $F_{\mathrm{producer}}$ retains at most three semantically relevant producers. Retained producers become new anchors; expansion stops when no unseen anchor remains. The reached nodes and producer-to-consumer edges form the relevant information graph $\mathcal{G}_{\mathrm{info}}=(\mathcal{V}_{\mathrm{info}},\mathcal{E}_{\mathrm{info}})$. Although traversal is backward, edge orientation preserves the true producer$\rightarrow$consumer direction.

\begin{algorithm}[t]
\caption{Backward Relevance Tracing}
\label{alg:supp-backward-tracing}
\begin{algorithmic}[1]
\REQUIRE Query $Q$, trajectory $\tau$, structure $\mathcal{H}$, final failure step $x_{\mathrm{final}}$
\ENSURE Relevant information graph $\mathcal{G}_{\mathrm{info}}$
\STATE $\mathcal{A}\leftarrow\{x_{\mathrm{final}}\}$; $\mathrm{Visited}\leftarrow\emptyset$
\STATE $\mathcal{V}_{\mathrm{info}}\leftarrow\{x_{\mathrm{final}}\}$; $\mathcal{E}_{\mathrm{info}}\leftarrow\emptyset$
\WHILE{$\mathcal{A}\neq\emptyset$}
    \STATE Pop an anchor $x_j$ from $\mathcal{A}$
    \IF{$x_j\in\mathrm{Visited}$} \STATE \textbf{continue} \ENDIF
    \STATE $\mathrm{Visited}\leftarrow\mathrm{Visited}\cup\{x_j\}$
    \STATE $\mathcal{P}_{\mathrm{intra}}\leftarrow$ intra-subtask predecessors of $x_j$
    \STATE $\mathcal{P}_{\mathrm{inter}}\leftarrow$ cross-subtask producers matched by data item
    \STATE $\widehat{\mathcal{P}}_{\mathrm{inter}}\leftarrow F_{\mathrm{producer}}(Q,x_j,\mathcal{P}_{\mathrm{inter}})$
    \FOR{each $x_i\in\mathcal{P}_{\mathrm{intra}}\cup\widehat{\mathcal{P}}_{\mathrm{inter}}$}
        \STATE Add $x_i$ and $(x_i\!\rightarrow\!x_j)$ to $\mathcal{G}_{\mathrm{info}}$
        \STATE Add unseen cross-subtask producer $x_i$ to $\mathcal{A}$
    \ENDFOR
\ENDWHILE
\RETURN $\mathcal{G}_{\mathrm{info}}$
\end{algorithmic}
\end{algorithm}

\paragraph{Forward annotation roles.}
EMFA then processes $\mathcal{G}_{\mathrm{info}}$ in chronological order and attaches step-level annotations $\Phi_{\mathrm{err}}$, yielding $\mathcal{G}_{\mathrm{prop}}=(\mathcal{G}_{\mathrm{info}},\Phi_{\mathrm{err}})$. The formal role set used in the main paper is:
\emph{propagation source},
\emph{propagation carrier},
\emph{recovered deviation},
\emph{unrecovered deviation},
and \emph{failure symptom} (exposed at $x_{\mathrm{final}}$).
In implementation, finer diagnostic cues---such as method or query framing, execution-plan authorship, information production, and acceptance or transformation boundaries---are mapped onto these roles. Early irregularities that initiate a reconstructed path are labeled as the propagation source; intermediate nodes that transfer or transform erroneous information are propagation carriers; irregularities later corrected are recovered deviations; those that remain active and affect later decisions are unrecovered deviations; and the terminal exposure of the observed failure is the failure symptom. A propagation source need not coincide with the decisive error returned by failure attribution. Simple deterministic checks for explicit or omitted required actions complement the LLM annotations.

\subsection{B.3~Loop Extraction}

Producer--consumer propagation does not capture failures in which agents repeatedly pursue the same unresolved objective without substantive progress. EMFA therefore extracts persistent interaction loops independently of $\mathcal{G}_{\mathrm{info}}$.

\paragraph{Loop-group recovery.}
An LLM compares objectives, actions, observations, and progress across repeated interaction rounds in $\mathcal{H}$ and groups steps that address the same unresolved objective into loop groups. The resulting $\mathcal{G}_{\mathrm{loop}}$ connects those steps even when they span multiple subtasks or do not follow an ordinary information-dependency path.

\paragraph{Loop-level candidates and boundary rules.}
Each loop group is analyzed to determine whether repetition originates from an ambiguous or incorrect planning instruction, or from an executor that fails to follow or correctly execute a valid instruction. The first occurrence of the same unresolved failure category is selected as the group-level candidate; an aggregate selector retains at most two loop candidates in $\mathcal{X}_{\mathrm{loop}}$. Boundary adjudication used later in counterfactual verification checks recurring multi-agent patterns---including planner--executor responsibility, code--runtime boundaries, evidence--verification boundaries, and premature declarations of completion---so that responsibility is not assigned to a downstream surface actor when the decisive error is introduced upstream.

%% file: Supplement/Content/AdditionalResults.tex
\section{C.~Additional Experiments}

\subsection{C.1~Cost Analysis}

\input{Content/table/token_cost}

To quantify the inference overhead of propagation-aware attribution, we compare average token consumption per failure log across representative methods. As shown in Table~\ref{tab:supp-token-cost}, EMFA consumes more tokens than existing prompting and structure-guided approaches, requiring approximately $3.1\times$--$3.5\times$ the tokens of CHIEF on the two Who\&When subsets. The additional cost comes from structured trajectory construction, dual-branch error-propagation modeling (including forward annotation and loop extraction), candidate screening, and counterfactual verification. Unlike replay-based attribution, however, EMFA does not require repeated environment executions: the increased inference budget purchases substantially stronger step-level attribution without interactive rollouts.

%% file: Supplement/Content/table/token_cost.tex
\begin{table}[t]
    \centering
    \setlength{\tabcolsep}{4pt}
    \caption{Average token cost per case on Who\&When. Lower is better.}
    \label{tab:supp-token-cost}
    \begin{tabular}{@{}lcc@{}}
    \toprule
    \multirow{2}{*}{\textbf{Method}} & \multicolumn{2}{c}{\textbf{Token Cost Per Case} $\downarrow$} \\
    \cmidrule(lr){2-3}
     & Hand-Crafted & Alg.-Generated \\
    \midrule
    All-at-Once & 21,581 & 5,833 \\
    Step-by-Step & 87,720 & 6,533 \\
    Binary Search & 34,659 & 5,226 \\
    ECHO & 53,701 & 25,642 \\
    CHIEF & 55,085 & 19,504 \\
    \textbf{EMFA} & \textbf{195,145} & \textbf{60,439} \\
    \bottomrule
    \end{tabular}
\end{table}

%% file: Supplement/Content/CaseStudy.tex
\section{D.~Additional Case Studies}

We present a Hand-Crafted case deferred from the main paper. Figure~\ref{fig:supp-case-study} visualizes the trajectory and the error-propagation graph reconstructed by EMFA.

The query is: ``What is the last word before the second chorus of the King of Pop's fifth single from his sixth studio album?'' The agents correctly identify \textit{Thriller} as Michael Jackson's sixth studio album (steps~3--4) and \textit{Human Nature} as its fifth single (steps~6 and~8), then request a lyric lookup (step~10). The ground-truth decisive error is at step~12, where the lyrics search fails to provide the correct OCR text. Contaminated lyrics then propagate through subsequent analysis (step~14), leading the system to infer ``bite'' (step~16), confirm it in the ledger (step~17), and return the wrong answer (step~19).

The Step-by-Step baseline incorrectly predicts step~16---a propagation carrier induced by corrupted evidence---as the decisive error. By constructing $\mathcal{G}_{\mathrm{prop}}$ from $x_{\mathrm{final}}$, screening dual-branch candidates, and applying counterfactual verification, EMFA localizes the decisive error to step~12, disentangling it from both downstream carriers and the failure symptom exposed at the final failure step.

\begin{figure*}[t]
\centering
\includegraphics[width=\textwidth]{{../Figures/case_study}.png}
\caption{
Case study of EMFA's error-propagation analysis on a Hand-Crafted trajectory.
\textbf{Left:} Original trajectory with reasoning-irrelevant steps omitted.
\textbf{Right:} Error-propagation graph constructed by EMFA, showing how the decisive error emerges, propagates, and culminates in task failure.
The \textit{Baseline-Predicted Error} denotes the Step-by-Step prediction, which localizes to a propagation carrier rather than the decisive error.
}
\label{fig:supp-case-study}
\end{figure*}

%% file: Supplement/Content/RepresentativePrompts.tex
\section{E.~Representative Prompts}

We summarize the key prompt families used by EMFA, aligned with the five stages in the main paper. Full templates are released with the code; here we retain only the decision constraints that define each stage.

\paragraph{Final Failure Step Identification.}
The model first extracts verifiable success criteria and hard constraints from the query alone, without solving the task or guessing the answer (JSON fields such as \texttt{success\_criteria} and \texttt{constraints}). Given the last $w{=}5$ trajectory steps, it then judges whether a step states an explicit final answer; the selected step is compared against $C_{\mathrm{success}}$ to characterize the failure symptom at $x_{\mathrm{final}}$.

\paragraph{Structured Trajectory Construction.}
The model partitions the log into contiguous subtasks and recovers producer--consumer dependencies. A representative system instruction is:
\begin{quote}
\small
\textit{You are analyzing a multi-agent execution log. Decompose the trajectory into contiguous semantic subtasks and recover information dependencies. Do not answer the user query. Return strict JSON only.}
\end{quote}

\paragraph{Error Propagation Modeling.}
Prompts supply $\mathcal{G}_{\mathrm{info}}$ or loop-group context and ask the model to annotate introduction, propagation, transformation, or correction of deviations along the reconstructed failure-related structures $\mathcal{G}_{\mathrm{prop}}$ and $\mathcal{G}_{\mathrm{loop}}$.

\paragraph{Dual-Branch Candidate Screening.}
Given the annotated propagation and loop structures, the model selects at most two candidates per branch, preferring the earliest unrecovered deviation that materially affects later actions, and forms the deduplicated pool $\mathcal{X}_{\mathrm{cand}}$.

\paragraph{Counterfactual Verification.}
For each candidate $(a_i,i)\in\mathcal{X}_{\mathrm{cand}}$, the verifier considers a minimally repaired version of that step and assesses whether the repair would block the downstream failure process and satisfy $C_{\mathrm{success}}$. Remaining ties are resolved by responsibility-boundary principles (planner vs.\ executor, transformation vs.\ raw evidence, premature completion vs.\ earlier limitation).